\documentclass[letterpaper, 10 pt, conference]{ieeeconf}  

\IEEEoverridecommandlockouts                              

\makeatletter
\let\NAT@parse\undefined
\makeatother

\usepackage[numbers,sort&compress]{natbib}

\usepackage[utf8]{inputenc}
\usepackage{graphicx}
\usepackage{amsmath}
\usepackage{amssymb}
\usepackage{tabularx}
\usepackage{caption}
\usepackage{xurl}
\usepackage[T1]{fontenc}

\DeclareCaptionLabelSeparator{periodspace}{. \ } 
\title{\LARGE \bf
Teaching is a Process: The TOSS Framework for Modeling Human Teaching Decisions in Human-Interactive Robot Learning
}

\author{Bernhard Hilpert$^{1, 2}$, Kim Baraka$^{2}$ and Joost Broekens$^{1}$
\thanks{$^{1}$LIACS,
         Leiden University, Leiden, The Netherlands
         }%
\thanks{$^{2}$
         VU Amsterdam, Amsterdam, The Netherlands
         }%
}

\begin{document}

\maketitle
\thispagestyle{empty}
\pagestyle{empty}

\begin{abstract}

 Successful Human-Robot Teaching assumes alignment between robot processing needs and human teaching intent. 
    To better understand this alignment, this work seeks to uncover the underlying logic that humans intuitively apply when teaching.
    Through an exploratory, bottom-up study with \textit{N}=34, participants observing two distinct robot Reinforcement Learning (RL) scenarios, we analyze 204 intuitive teaching responses across early, middle, and late learning phases.
    Results reveal that teaching decisions consist of a nuanced, interconnected network of Triggers (situational catalysts), Objectives (subjective teaching targets), Signals (communicative acts), and Strategies (high-level governance) in which teachers spontaneously adopt diverse roles, acting as coaches, engineers, or designers and prioritize different objectives. 
    Based on these results, we introduce the TOSS Framework, which conceptualizes Human-Robot teaching as a procedural loop between robot behavior and human teaching actions, in which human teaching decisions are modeled as Trigger-Signal responses modulated by teaching Objectives and Strategies. It provides future research with an openly accessible dataset and a theoretical foundation for a) understanding teaching decisions and b) simulating realistic oracles as well as c) designing human-centered teaching settings and novel robot learning algorithms that go beyond the constraints of current robot learning settings.
\end{abstract}

\section{Introduction}

In Human-Interactive Robot Learning (HIRL) \cite{baraka_humaninteractiverobotlearning_2026}, human teachers are integrated into the robot learning process through interactive feedback cycles to achieve Hybrid Intelligence \cite{akataResearchAgendaHybrid2020} by assisting robot learners on tasks they cannot master alone. 
However, current HIRL paradigms often force teachers into reactive roles within high-pressure learning loops, failing to account for the procedural logic humans naturally apply when educating a peer \cite{chetouaniInteractiveRobotLearning2023}, often resulting in suboptimal teaching behavior \cite{gao_reinforcementlearningimperfect_2019}. 
Research has attempted to bridge this gap by analyzing after-the-fact justifications of teaching \cite{hoPeopleTeachRewards2019, ramaraj_understandingintentionshuman_2020} and teacher mental models \cite{richterImprovingHumanRobotTeaching2025, ramaraj_unpackinghumanteachers_2021} or inferring teaching goals algorithmically \cite{tulliInferringImplicitGoals2025}. 
While these are vital steps towards ``opening the human black box", they remain constrained by the very interactive loops they seek to explain: 
As teachers observe robot errors, they are forced to strategically adapt their feedback to compensate for the robot’s procedural flaws \cite{huangEffectRobotErrors2024}.
Consequently, what is observed in these settings is not a pure reflection of the teacher's intuitive teaching approach, but rather a strategic ``mask" adapted to a misaligned teaching system.
This creates a fundamental gap: research lacks a baseline understanding of the ``unfiltered" pedagogical logic of teaching decisions, humans would naturally employ if they were not forced into constant strategic compensation. 
To uncover this baseline, this work decouples the teacher from the immediate pressures of the learning loop. 
By employing an unconstrained, non-interactive experiment with three measurement intervals over two different Reinforcement Learning (RL) settings (tabular Q-learning in a navigation task and a Deep Deterministic Policy Gradient (DDPG) agent in a manipulation task, as a proxy for other HIRL algorithms), this work moves beyond the analysis of reactive adaptation. 
It provides a bottom-up, inductive description of raw triggers, objectives, signals and strategies that constitute human teaching decisions made available as an open-access dataset and introduces the integrated TOSS framework as a characterization of the procedural logic humans intuitively gravitate towards when teaching robot learners. 
This framework can be applied in HIRL for modeling realistic oracles to evaluate algorithms through simulated teachers, an empirically grounded interpretation of human teaching decisions, and teacher-centered redesign of robot teaching settings. 

\section{Related Work \& Background}
\label{Related Work}

\label{Background}
The integration of targeted human feedback into the robot learning process can improve learning speed and adaptation to novel settings by formalizing human teaching signals such as Feedback, Demonstration and Instruction in a robot-understandable way \cite{chetouaniInteractiveRobotLearning2023}.
However, the requirements of this formalization process for the teaching set-up are not always ``human-friendly": cumbersome teaching settings can lead to fatigue and the loss of focus 
or motivation \cite{guillorySimultaneousLearningCovering2011}.
As a result, human feedback is often suboptimal or ``noisy" \cite{houShapingImbalanceBalance2023, gao_reinforcementlearningimperfect_2019}.
Consequently, improved algorithmic interpretation of ``richer" human feedback signals has been developed, i.e., learning from inaccurate feedback \cite{faulknerInteractiveReinforcementLearning2020}, sample-efficient active learning \cite{houGiveMeExample2024} 
and robot transparency mechanisms \cite{habibianReviewCommunicatingRobot2023}, including the recent use of Large Language Models \cite{richterImprovingHumanRobotTeaching2025}. 
This essentially opens the ``AI blackbox" and 
makes the procedural functioning of robots more accessible.
However, all of these approaches treat human variability as a technical hurdle to be overcome.

A consistent line of research analyzing (flawed) teaching behavior shows that human teaching is often a communicative act: teachers 
use feedback signals as communication about desired behavior, rather than rewards \cite{hoPeopleTeachRewards2019,sarinPunishmentOrganizedPrinciples2021}. 
While this reveals a fundamental misalignment between human teaching intent and the procedural functioning of robot learners, recent work suggests that this misalignment is systemic rather than random \cite{carrollUtilityLearningHumans2019}. 
Consequently, the variability (or noise) in human teaching is better understood as a feature that provides insight into the teacher's internal mental model of the learning process \cite{huangLeveragingVariationHuman2025}. 
Mental models are internal representations that humans use to interact with the world around them \cite{villareale_understandingmentalmodels_2021}.
However, efficient knowledge transfer requires the matching of the teachers' mental model with the capabilities of the robot learner \cite{bansalAccuracyRoleMental2019}. 
Current research on mental model mismatches (MMMs) clarified that they seem to be rooted in teacher intentions mismatching the robot learning process \cite{richterImprovingHumanRobotTeaching2025}.  
\cite{ramaraj_understandingintentionshuman_2020} used a theory-driven, deductive approach based on collaborative discourse theory to successfully taxonomize the semantic layer of teacher intentions.  
However, while mental model research focuses mostly on high-level conceptual models for semantic alignment on a task level, technical research focuses primarily on technical explainability for algorithmic alignment on a procedural level, creating a disconnect between the levels of analysis \cite{habibianReviewCommunicatingRobot2023}. 
Evidence for this procedural disconnect is quantified in findings that correlate persisting teaching patterns with both robot errors \cite{huangEffectRobotErrors2024} and human attributes \cite{fangCHARMConsideringHuman2025}. 

While recent efforts have begun to model this variation in feedback expression in order to enable more natural teaching \cite{huangLeveragingVariationHuman2025}, 
these teacher models are still derived from interactive settings.
In such settings, teaching behavior is highly volatile, suggesting that teachers are forced to deviate from their preferred strategies to compensate for the misalignment between their intuitive teaching policy and the robot’s inferred learning mechanics. 
Effectively, teachers `hack' their own signals to bridge this procedural gap \cite{huangEffectRobotErrors2024}. 
The patterns observed in such settings are thus not a pure reflection of the teachers' intuitive procedural approach to teaching, but rather a strategic adaptation to a misaligned teaching setting and an imperfect robot learner \cite{bansalAccuracyRoleMental2019}, which obstructs an unbiased examination of teachers' pedagogical logic, especially at later learning stages \cite{villareale_understandingmentalmodels_2021}. 
Consequently, the interactive nature of existing research produces results that function as a ``Reactive Mask" over the teacher's intuitive internal logic. 

Current understanding of teacher decisions therefore remains restricted to either theory-driven, deductive taxonomies or reactive post-hoc analyses of teacher adaptations to robot behavior. 
It is unclear what teachers intuitively prioritize when not forced into strategic compensation.
 To uncover the baseline intuitive teaching process, it is necessary to decouple the teacher from the immediate pressures and loop dependencies of the interaction. 
Thus, the goal of this study is to examine the unbiased teaching decisions, teachers intuitively gravitate towards, when observing robot learners.

\section{Methodology}
In this bottom-up study, an observational experimental paradigm was implemented.
Following recommendations for open research practices \cite{wesslerEmpiricalResearchAffective2021}, this experiment was preregistered, IRB-approved and all materials (Questionnaire, coding plan, stimuli, data corpus, preregistration protocols) are openly available 
on OSF\footnote{\url{https://osf.io/fumd8/?view_only=9cec60dccbd446f08bd818d0b3612705}}.

\subsubsection{Participants} A total of 35 participants was recruited through Prolific (1 excluded for inattentive answers), resulting in a final sample of \textit{N} = 34 (26 Male, 8 Female, 0 Non-binary, self-assigned or undisclosed), between 19 and 65 years old (\textit{M}\,=\,38.79 years, \textit{SD}\,=\,10.68) and received on average £10.42 GBP. 
Screening criteria included being above legal age, a 95\% approval rate or higher on Prolific (minimum of 100 completed tasks), and self-reported fluency in English with a residence 
in high English fluency countries.

\begin{figure}[h]
\centering
\vspace{-0.3cm}
  \includegraphics[width=\columnwidth]{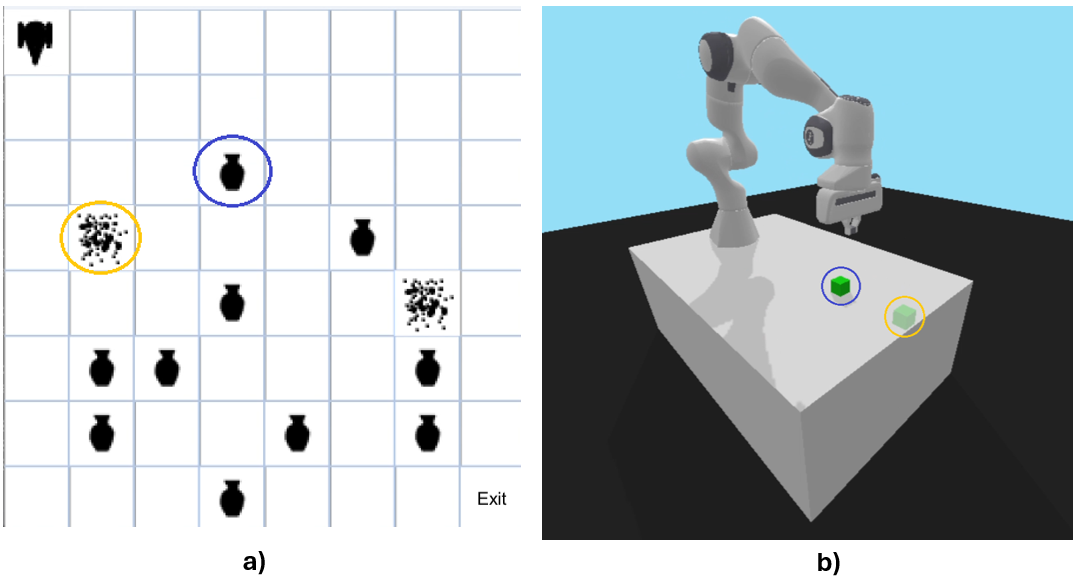}
  \caption{a) NT: Cleaning robot: dirt to clean (yellow), furniture to avoid (blue) b) MT: Assistive robot: medication (blue) to push to target (yellow)}
  \label{fig:Tasks}
  \vspace{-0.4cm}
\end{figure}

\subsubsection{Experimental set-up} In line with previous work \cite{ramaraj_understandingintentionshuman_2020}, 
a qualitative questionnaire study was employed in this experiment.
Participants observed a RL robot's learning process with a framing of being a teacher but without the possibility of intervention. 
The questionnaire started with a procedure briefing, an informed consent form and a commitment request.
Then, they received an introductory description 
of HIRL in lay user terms\footnote{Instructions, questionnaire and stimuli were piloted with \textit{N=}4 test participants to ensure understandability}, before the data collection started.
The assessment presented participants with two scenarios (see Section \ref{Scenarios} below) in which a robot learned to solve a task. The stimulus videos for each task (see Section \ref{Stimuli}) were split-up in 3 parts of equal length representing early, middle and late training phases (6 videos in total) to account for the dynamically progressing nature of RL. 
For each video, participants' teaching intuitions were assessed in a free-form format with the question ``\textit{If you could, how would you help the robot in this video to learn the task better?} to capture their intuitive teaching approach.\footnote{The data collection in this experiment was part of a larger survey that also examined teacher observations, reported in \cite{hilpert_canyousee_2025}}
After completion (median: 57 minutes), they were thanked and redirected to Prolific.

\subsubsection{Scenarios}
\label{Scenarios}
In order to ensure generalizability of the results, the experiment presented participants with two robot embodiments learning \emph{two different tasks}: one learned a grid-world navigation task (NT) and the other a manipulation task (MT) with a robotic arm.
This serves to test whether teaching intuitions in robot learning apply to both tabular and function approximation learning or if there are differences.
The first study utilized a tabular Q-learning agent performing an NT in an 8x8 grid-world environment (based on \cite{brockman_openaigym_2016}). To foster a realistic teaching context and mitigate anthropomorphic bias, the task was framed as a domestic cleaning scenario (see Fig. \ref{fig:Tasks}): white tiles represented floor space, iconized vases represented furniture, and the goal was marked as exit. To increase the face validity and make learning progress more visible to observers, two dynamic `dirt patches' were included as intermediate milestones, which granted a one-time reward (+1, equivalent to the exit reward) and disappeared upon contact. Participants observed rendered recordings of the robot's learning process, with robot depicted as a conic cleaning robot to clearly indicate its orientation.
The second agent was a Deep Deterministic Policy Gradient (DDPG, an off-policy, model-free, actor-critic) algorithm 
performing a continuous MT.
This task was framed as an assistive robot in elderly care learning to push a `medication box' across a table to an accessible target for a patient. Stimuli consisted of rendered recordings of the robot in a modified PandaGym `Push' environment \cite{gallouedecPandagymOpensourceGoalconditioned2021}, with a fixed object starting and a target position closer to the table's edge. The agent's reward function was a composite of the distances between the end-effector, the object, and the target, with a sparse +1000 bonus upon successful object-target contact.

\subsubsection{Stimuli}
\label{Stimuli}
To reflect a realistic learning progression, stimuli were rendered from trajectories at regular intervals during training. Convergence was defined as five consecutive successful task completions. For the NT, checkpoint policies were saved every 500 episodes over a 10000-episode training run, resulting in a 94-second stimulus video. For the MT, checkpoints were saved every 1000 steps over 100000 training steps, totaling 192 seconds of stimulus video footage. 

\subsubsection{Analysis}
To extract insights without theoretical bias, an inductive, reflexive thematic analysis was conducted based on \cite{braun_thematicanalysispractical_2021}, following a three-phase pipeline: (1) Data Reduction: condensing statements into core semantic units and grouping by synonymy; (2) Theme Generation: clustering core meanings into latent themes with descriptive labels; and (3) Refinement: cross validating themes against the full dataset to ensure internal consistency.
To maintain conceptual continuity across datasets, primary coding was executed by a single investigator, following \cite{braun_thematicanalysispractical_2021}. To establish systemic trustworthiness and counteract individual coder bias, the evolving thematic structure was subjected to 30 hours of systematic peer-debriefing and critical review with senior HRI researchers, followed by an external validation workshop ($N=26$) to verify the accuracy of the generated framework.

\section{Results}
\label{Results}

The collected data showed a large heterogeneity in the teacher statements. 
The thematic analysis revealed that  - when unbiased by instruction on how and when to intervene - 
teachers intuitively conceptualize teaching not as a monolith, but rather as a nuanced, multi-faceted process, consisting of: 

\begin{enumerate}
\item Triggers: situational catalysts for intervention
\item Objectives: subjective targets or desired goal behaviors
\item Signals: communicative acts or teaching methods
\item Strategies: high-level governance of the teaching sequence
\end{enumerate}

\subsection{Triggers: The ``Why" of Teaching}
\label{Triggers}
Triggers are defined as observed situated robot behaviors that prompt the teaching approach. 
They serve as the ``gateway" to the teaching process by identifying moments where a teacher’s subjective mental model of the task or robot conflicts with observed behavior (rather than representing objective robot states).
The data suggests that teachers evaluate robot behavior along three continuous criteria:

\textbf{(A) }\underline{\textit{Consistency (Systematic vs. Incidental):}} Is it evaluated as a persistent pattern (S) or a discrete event (I)?

This focuses on the perceived frequency of the behavior.
Systematic Triggers identify perceived patterns in the robot's behavior. Here, the teacher perceives a recurring trend in the current robot behavior (\textit{``it is overshooting a lot currently"}).
Incidental Triggers react to discrete or intermittent events (``glitches"). Here, the teacher identifies a moment of unpredictability that is perceived as an outlier in the robot’s current learning (\textit{``it sometimes moved dramatically horizontally"}).

\textbf{(B) }\underline{\textit{Reference Frame (Relative vs. Absolute):}} Is the behavior evaluated against the teacher's subjective model of the \textit{robot's progress} (R) or against their understanding of the \textit{task's objective logic} (A)?

This focuses on the context of evaluation. 
Relative Triggers evaluate behavior against the \textit{observed past or anticipated future behavior}. Here, the robot might be technically suboptimal, but closer to optimal than it was earlier (\textit{``the robots aim has improved a lot"}).
Absolute Triggers evaluate behavior against the teacher's understanding of the \textit{task}. Here, the robot's behavior might go against the teacher's understanding of the requirements for task completion, regardless of the robot's previous or expected future performance (\textit{``it fails to even touch the box"}).

\textbf{(C) }\underline{\textit{Alignment (Desired vs. Undesired):}} Does it align with (D) or violate (U) the teacher's goals for the robot's learning? 

This focuses on the goal-congruence of the behavior.
Undesired Triggers identify behaviors that violate the teacher’s goals for the task or the robot’s learning (\textit{``Its movements were too erratic"}). Here, the teacher identifies a shortcoming or error that necessitates a corrective or preventive intervention.
Desired Triggers identify behaviors that align with the teacher’s goals for the task or the robot’s learning (\textit{``it's doing very well"}). Here, the teacher identifies a moment of progress that warrants reinforcement.

These evaluations intersect and crystallize into distinct trigger clusters which represent how teachers categorize robot behavior before intervening. 
A selection of these clusters, illustrated through thematic anchors that define the boundaries of each dimension, is presented in Table \ref{tab:Triggers}.

\begin{table}[h]
\caption{Overview of Triggers: Clusters Represent the thematic anchors at the boundary intersections of evaluative criteria (see \ref{Triggers} A-C)}
\label{tab:Triggers}
\begin{tabularx}{\columnwidth}{|p{0.75cm}|p{2.1cm}|X|}
\hline
\textbf{Cluster} & \textbf{Label} & \textbf{Example} \\
\hline
SAU & Consistent Failure & \textit{``it seems to wonder around aimlessly until it stumbles into dirt"} \\
\hline
IAU & Incidental Failure & \textit{``at times it is completely off the table"} \\
\hline
IRU & Incidental Suboptimal Choice & \textit{``take[s] unnecessary steps which causes the vases to break"} \\
\hline
SRU & Suboptimal Behavior Pattern & \textit{``avoid tiles it has already visited"} \\
\hline
IRD & Isolated Progress & \textit{``To know that once it has located the box then just to go forward"} \\
\hline
SRD & Performance Improvement & \textit{``really improved its pathway and getting to the exit"}  \\
\hline
IAD & Incidental Correct Behavior & \textit{``give positive feedback when the robot orients towards the mess"} \\
\hline
SAD & Mastering & \textit{``I think the robot has it all"} \\
\hline
\end{tabularx}
\vspace{-0.7cm}
\end{table}

\subsection{Objectives: The ``What" of Teaching}
Objectives are defined as subjective target states and represent the specific goal behaviors, teachers intend to achieve with their intervention.
The data suggests that teachers intuitively conceptualize objectives as subjective targets for three distinct functional levels of the robot: 

\textbf{(A) }\underline{\textit{Behavioral Objectives}} represent subjective targets related to the process of task execution. They are essentially target modifications of current behavior that teachers would like to achieve through their teaching independently of task outcome. If the robot achieves the goal but does so ``messily", a Behavioral Objective is activated. This includes the systematism (\textit{``to work horizontally or vertically"}), precision (\textit{``I would aim for it to slow down a bit"}), efficiency (\textit{``not go back over areas its seen as safe"}) or stability (\textit{``repeating the process many times"}) of behavior.

\textbf{(B) }\underline{\textit{Goal/Task Objectives}} represent subjective targets related to task achievements and essentially focus on specific milestones and task outcomes that the teacher would like to achieve through their teaching independently of execution. If the robot shows convergence-like, systematic behavior but misses out on collecting rewards, a Goal Objective is activated. This includes mastering a subtask (\textit{``concentrate more on grasping the medication before anything else"}) , task decomposition (\textit{``seek out the closest patch first, then, move onto the furthest patch and then exit"}), and preventing mistakes (``\textit{avoid damaging furniture"}). 

\textbf{(C) }\underline{\textit{Knowledge Objectives}} represent subjective targets related to the robot's internal model or an understanding of the environment and the task. This includes environmental mapping (\textit{``give it the ability to map the room"}), knowledge of object location (\textit{``let the robot know where the dirt's and vase are"}) as well as functional declarations that communicate the purpose or identity of objects (\textit{``i'd make the medication and the target clearer"}) and goals (\textit{``Give it a goal"}) to the robot.

\subsection{Signals: The ``How" of Teaching}
\label{Signals}
Signals encompass the method of how a robot should be teached and represent the interactive part of the teaching process that defines the actual intervention. 
Five different clusters of intuitive teaching signals emerged from the data:

\textbf{(A) }\underline{\textit{Evaluative Feedback}} refers to positive or negative signals on past or (assumed) future behavior. Here, the teacher acts as a Judge that is (in-)validating robot behavior by reinforcing or punishing it.
This includes statements such as \textit{``give positive feedback when the robot orients towards the mess"} or \textit{``offering clearer rewards and penalties"}.

\textbf{(B) }\underline{\textit{Corrective Advice}} refers to signals that are relative adjustments of past or (assumed) future wrongful behavior. Here, the teacher acts as a Tutor that provides a specific modification relative to the robot's attempt. This includes statements such as \textit{``if it managed to get to the same place in the same alignment again, it would turn the opposite direction from the previous time"} or \textit{``slowing the arm down, maybe taking away its sideways movements and only allowing a back and forth action"}.

\textbf{(C) }\underline{\textit{Demonstration}} refers to guidance signals of target behavior. Here, the teacher acts as an Example for the robot to learn from by providing absolute reference trajectories. This includes statements such as \textit{``travel along one axis to the edge of the space then move down \& travel in the opposite direction \& repeat"} or \textit{``position the robot behind the bottle of pills and shove it backwards"}.

\textbf{(D) }\underline{\textit{Information Sharing}} refers to acts of communicating information about the environment or the task. Here, the teacher acts as an Oracle, providing declarative information about learning context or goal specifications. 
This includes \textit{``provide a map with dirty area and furniture"} or \textit{``Giving it a goal to get to an exit when no more dirt is present"}. 

\textbf{(E) }\underline{\textit{Withholding}} refers to the deliberate choice to refrain from interfering as a functional (null) signal (\textit{``I would make sure to not move anything around after it started to learn the optimal path."}). Here, the teacher acts as an Observer. This is not an absence of teaching, but a pedagogical decision to foster independence, or test the robot's learning convergence.

\subsection{Strategies: The Teaching Approach}
Strategies represent the high-level governance of the teaching process that allude to a general teaching policy.
They define the locus of intervention and strategic profile of the teacher in the interaction (as opposed to the tactical roles in \ref{Signals}) and within that, their teaching mode or plan for robot development.
The data suggests five different strategies: 

\textbf{(A) }\underline{\textit{Interactive Guidance}} represents real-time, ``in-the-loop" feedback cycles in which the teacher manages robot's learning dynamically by reacting to situational Triggers and combining them with specific Objectives and Signals in adequate sequences to form a cohesive interactive loop. Here, the teacher acts as a Coach.
This strategy is characterized by its varying depth, ranging from brief tactical nudges (e.g., \textit{``move slower"}) to comprehensive instructions that explain the entire reasoning chain behind a correction: \textit{``The robot should concentrate on using smaller movements to first locate the medicine, and then efficiently move the medicine to the desired location [...] I would encourage the robot to use more subtle and efficient movements"}.

\textbf{(B) }\underline{\textit{System Adaptation}} represents a ``meta" strategy where the teacher views the robot's current configuration as fundamentally insufficient. Here, the teacher acts as an Engineer, who rather than directly interacting with the robot, intends to enhance its fundamental architecture. This includes providing additional features like sensors (\textit{``a sensing or visual feature which allows it to detect large objects"}), tools (\textit{``Fix a clip-like hand [...] to grip the drug"}) or learning algorithms (``\textit{write a function that takes all possible degrees of freedom [...] 
and optimise it towards the best solution"} or \textit{``the robot needs to be programmed to learn from its mistakes"}).

\textbf{(C) }\underline{\textit{Task Adaptation}} represents a second ``meta" strategy of teaching by scaffolding or adapting the task. Here, the teacher acts as a Designer, using methods like adapting the environment (``\textit{i wouldnt have that many vases"}) or task complexity (\textit{``provide the robot with more diverse training scenarios"}) rather than the robot itself. The teaching happens by keeping the robot in a zone of proximal development and shifting that along as the teacher wishes. 

\textbf{(D) }\underline{\textit{Laissez-faire}} represents the deliberate choice to let the robot explore or self-learn without teacher interference. Here, the teacher acts as a non-interventionist to maximizes autonomous learning by removing human intervention. 

\textbf{(E) }\underline{\textit{Satisfied Supervision}}: represents the strategic decision in the special case that the robot has reached a perceived state of mastery. Unlike Laissez-faire, it reacts to perceived mastery rather than acting as an instructional strategy.

\subsection{The TOSS Framework: Teaching as an integrated process}

\begin{figure}[t]

\includegraphics[width=\columnwidth]{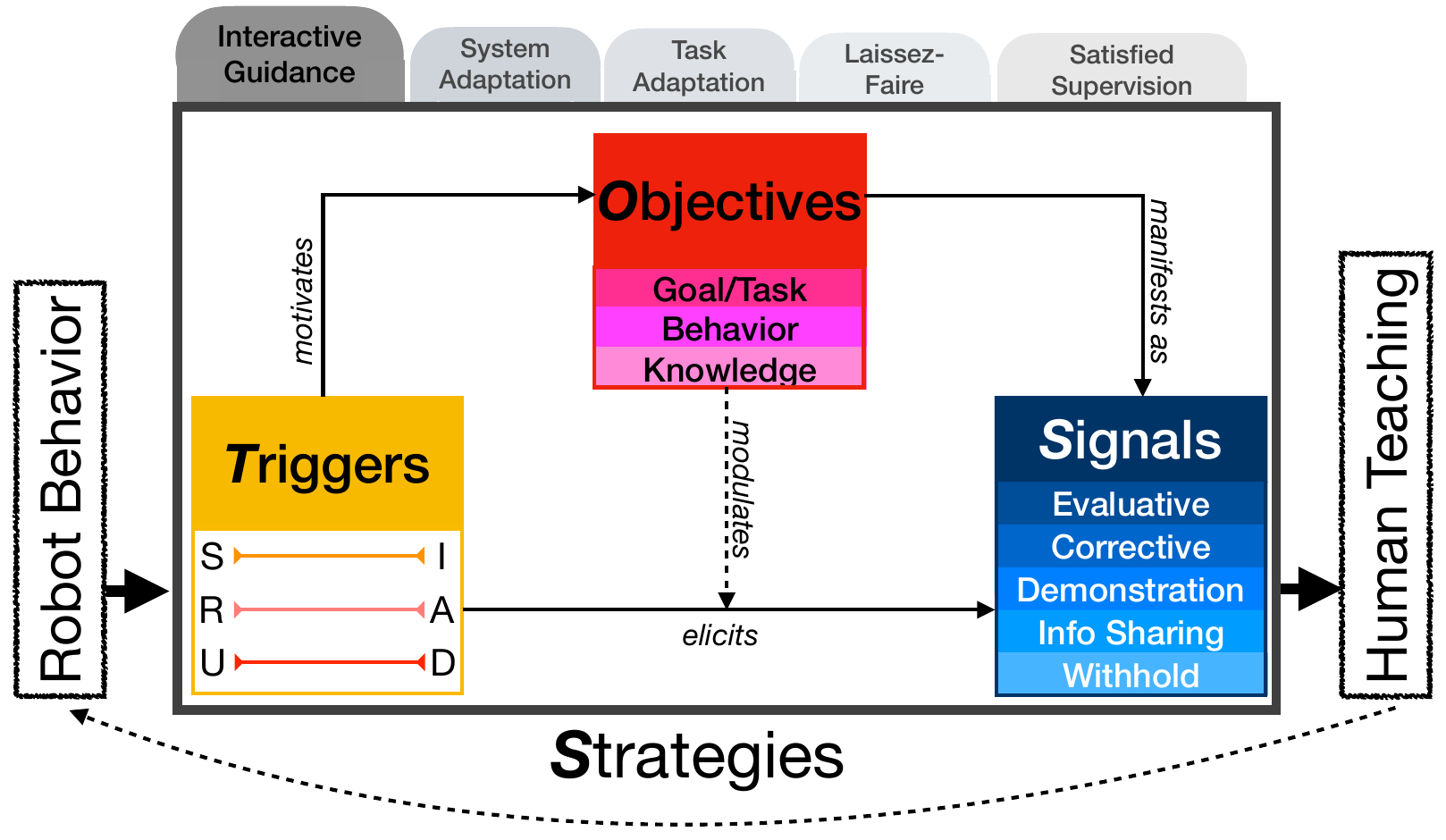}

\caption{The TOSS Framework: Arrows indicate the flow from the evaluation of observed robot behavior on three continuous dimensions (Triggers) to subjective targets (Objectives) and communicative acts (Signals), all modulated by high-level Strategies.}
\vspace{-0.65cm}

\label{fig:TOSS}

\end{figure}

While these results provide an analytical decomposition of teaching, the data reveals that these facets function as an integrated, multi-layer cognitive process. 
We formalize this process as the TOSS Framework (Triggers, Objectives, Signals, and Strategies, see Figure \ref{fig:TOSS}).

Rather than being a singular act, teaching emerges as a modular, iterative process where the different facets interlock to form a cohesive, continuous intervention.
In a complete cycle, an intervention is initiated by a Trigger (the perceived `why'), compared to an Objective (the mental `what'), manifested in teaching through a Signal (the communicative `how') - all nested within and governed by a high-level Strategy (the `identity' and plan of the teacher).
However, in the data the facets are often mixed within a single utterance.
For example: \textit{``I would encourage shorter, more precise movements because [...] it sometimes moved dramatically horizontally"} simultaneously encodes a Trigger (IAU), a Behavioral Objective (Precision), and a Signal (Corrective Advice) within an Interactive Guidance strategy.

A key finding is the structural flexibility of this process. 
The teaching cycle is not always exhaustive. Teachers frequently bypass specific facets depending on their strategic locus of intervention. For instance, in a proactive interactive guidance strategy, teachers often bypass the Trigger to prevent errors before they occur. 
Similarly, an Engineer (System Adaptation) might implement a sensor to reach a Knowledge Objective without a specific failure, while a Designer employing a Task Adaptation strategy might alter the environment without ever issuing a communicative Signal (\textit{``By having the medication in a fixed point at the start"}).

Overall, TOSS provides a framework to conceptualize human-robot teaching as more than just the transmission of feedback signals. It is a multi-layered process where the teacher's strategic profile shifts dynamically between Coach, Engineer, and Designer to react to their own subjective evaluations of the robot's current behavior and use communicated signals to shape it towards their own subjective targets.

\begin{figure*}[htbp]

\centering

\includegraphics[width=\textwidth]{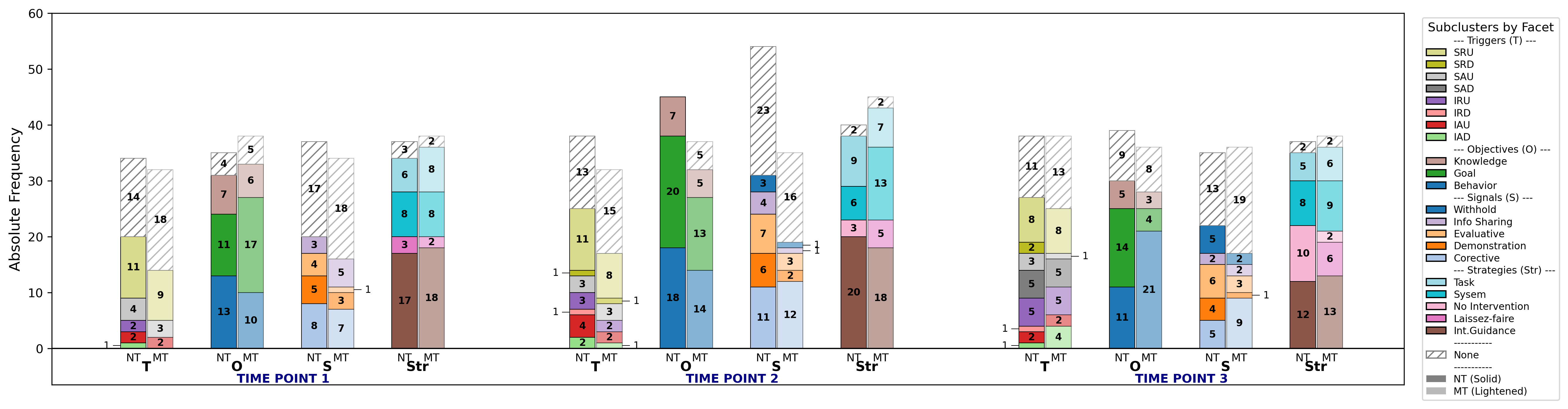}

\vspace{-5pt}
\caption{Distribution of TOSS-Components with respective Subclusters by Task and Time Point}
\vspace{-20pt}

\label{fig:Quantification}

\end{figure*}

\section{Discussion}

The results of this study complement current algorithm-focused research on teaching in HIRL by examining the pedagogical logic of the teaching decision process. 
This work presents the TOSS framework that conceptualizes intuitive teaching not as a monolithic act, but rather as an integrated, multi-faceted procedural loop where human mentors dynamically navigate between situationally adapted, high-level strategic profiles, continuously aligning subjective targets with subjective interpretations of shortcomings in a robot's learning by employing appropriate teaching signals.

\subsection{Triggers}
While existing literature on mental models in HRI acknowledges that human intervention is guided by more than simple task failure \cite{richterImprovingHumanRobotTeaching2025}, the TOSS framework demonstrates the intricate, multi-layered nature of intervention Triggers. 
A key finding is that Triggers function within a proactive internal evaluation process, rather than a mere reactive response to discrete behavioral errors.
This active role of the teacher may have been obscured in prior work by experimental designs that prompted specific intervention methods, thereby masking the teacher's spontaneous evaluative logic. 
When allowing for unconstrained teaching, the results reveal that human teachers do not simply scan for categorical, threshold-based ``right-or-wrong" errors. Instead, they maintain a state of continuous evaluation within three evaluative dimensions: Consistency, Reference Frame, and Alignment. 
The dimensional nature of Triggers suggests that teachers evaluate behavior through a sophisticated evaluative lens rather than a binary ``good or bad" scale. 
This suggests a significantly higher level of cognitive involvement than previously modeled.
By differentiating between Relative evaluations, where the robot is judged against its \textit{own behavior} and Absolute evaluations of \textit{task requirements}, these findings shed new light on the temporal and procedural dynamics through which teachers interpret robot behavior.
Similarly, the finding that teaching can both be informed by Consistent patterns or Incidental wrongdoings offers a more nuanced interpretation of how human teachers are triggered by or lenient towards ``errors" during the teaching process.
Collectively, these results suggest that Triggers should be seen in a much more dynamic and nuanced way than before. 

\subsection{Objectives}
A parallel interpretation can be applied to the teacher's conceptualization of teaching objectives. 
Prior research has either pre-defined objectives \cite{houProcessOrientedFrameworkRobot2023} 
or inferred them post-hoc from teacher behavior \cite{hoPeopleTeachRewards2019}. This work provides a direct query of the teacher’s own subjective targets.
Results show that teachers differentiate objectives into layers of behavioral execution- 
or goal-oriented objectives. 
The functional separation and the granular nature \textit{within} these targets indicate that human pedagogical intent operates at a much finer resolution than the high-level reward functions typically utilized. Furthermore, the findings suggest that objectives are not static: teachers may exhibit dynamic shifting between these levels (e.g. alternating behavioral and goal-oriented objectives throughout the teaching process), potentially even adjusting target granularity in response to the robot's perceived learning progress (i.e. shift from ``subtask mastering" to ``overall task achievement" towards the end, see Fig. \ref{fig:Quantification}). 
Most significant is the finding of Knowledge Objectives. While `information sharing' has been recognized as a communicative signal in oracle settings \cite{chetouaniInteractiveRobotLearning2023}, the identification of knowledge as an objective represents a critical departure from existing HIRL literature.
Particularly, it suggests an underlying human assumption that the robot possesses an internal world model capable of integrating declarative information into their learning.
This high level of abstraction in human thinking complements literature on MMMs  \cite{richterImprovingHumanRobotTeaching2025}.

\subsection{Signals}
The teaching signals largely mirror the taxonomies established in current human-robot interaction literature (i.e., evaluative feedback, corrective guidance, and demonstrations) that also formally map to RL \cite{chetouaniInteractiveRobotLearning2023, baraka_humaninteractiverobotlearning_2026}. 
While Information Sharing has been previously described in oracle settings \cite{chernova_robotlearninghuman_2022}, 
 the emergence of Withholding signals mirrors what is technically implemented as silent feedback or convergence testing \cite{loftinLearningBehaviorsHumandelivered2016}. 
The significance of these findings lies in their intuitive emergence. While previous work often predefines these signal channels for the user, this study demonstrates that lay teachers spontaneously converge on these signals even in unconstrained settings, which serves as a powerful validation of the current technical focus in HIRL as intuitive pedagogical rather than researcher-imposed constructs.
However, when viewed as a component within the TOSS framework, these signals should no longer be seen as simple isolated communicative acts. Instead, they can be seen as connectors between a diagnosed Trigger and a subjective target that the teacher wants to achieve. 
A trigger of Systematic-Absolute-Undesired (SAU) behavior (i.e., aimless wandering) diagnoses a fundamental lack of direction, which logically necessitates a Knowledge (defining the goal) or Goal Objective (Subtask Mastering) expressed through an Information Sharing or Demonstration signal.
Conversely, a trigger of Systematic-Relative-Undesired (SRU) behavior identifies a precision gap, which motivates a Behavioral Objective, that can be manifested through Evaluative Feedback or Corrective Advice.
By putting these standard signals in a broader procedural context, it becomes possible to interpret not just what was communicated, but why a specific signal seemed most fitting to satisfy a given strategic role.

\subsection{Strategies}
The distinct Strategies suggest that human teaching decisions are fundamentally tied to the teacher’s perception of their own strategic identity or profile within the interaction. 
The identification of these strategies necessitates a re-evaluation of the T-O-S (Trigger, Objective, Signal) sequence as a flexible, interactive process. Rather than a linear, rigid sequence, teaching should be seen in a broader temporal context of dynamic iteration.
Within that, the Laissez-faire strategy is not a mere absence of interaction. Instead, it represents a deliberate pedagogical choice to prioritize autonomous exploration that can represent an expectation of learning progress (``by now you should know it") or an overwhelmed dodging of responsibility  (``I have reached the end of my wisdom").
The emergence of Strategies highlights a critical teaching dynamic: teachers may provide interactive feedback up to a perceived capability ceiling, at which point they `jump' to scaffolding strategies like altering the environment or the system architecture to facilitate progress.
This mirrors previous work, where participants showed a tendency to provide evaluative feedback earlier in their interactions compared to demonstrations \cite{christofiHumanTeachingPatterns2025}. This work demonstrates that these temporally shifting teaching patterns are not limited to signals only, but also emerge for the teaching approach in general (see Figure \ref{fig:Quantification}). 
These findings suggest that current HIRL interfaces may be fundamentally misaligned with human intent. If a teacher adopts an `Engineer' profile but is restricted to a `Coach' interface that only accepts interactive teaching signals, a significant pedagogical mismatch occurs. Acknowledging this strategic flexibility is essential for designing systems that allow humans to fluctuate between profiles, moving seamlessly from teaching the robot to re-architecting the learning task as the interaction evolves.
While previous overviews have described these profiles within interactive robot learning contexts before \cite{chetouaniInteractiveRobotLearning2023}, they have mostly been 
studied separately from each other. 
In contrast, 
a key finding of the TOSS framework seems to be that (when unbiased by a predefined teaching setting) teachers naturally adopt Engineer 
and Designer 
profiles in addition or alternation to being Interactive Coaches.

\subsection{The TOSS Framework}
In general, TOSS allows for a formalization of teaching as a modular, non-linear process rather than a linear, fixed sequence of reactive acts. 
Previous work stated that ``humans provide teaching signals with an intention" \cite{chetouaniInteractiveRobotLearning2023}. 
The TOSS framework expands this idea, by providing a more nuanced picture that also accounts for a high degree of structural flexibility and interdependence, 
as, for example, teachers frequently skip situational Triggers to provide preemptive instruction or environmental changes, aiming to prevent errors before they occur. This suggests that the teacher's subjective target can be activated by the overarching Strategy alone, independent of the robot’s immediate behavior. 
In addition, the framework accounts for the intentional omission of a Signal despite a clear diagnostic Trigger.
In both instances, the components are not ``missing". Rather, they are strategically suppressed to optimize the teaching outcome.
This complements recent work advocating the consideration of human attributes \cite{fangCHARMConsideringHuman2025} and variation in human-robot teaching \cite{huangLeveragingVariationHuman2025} by introducing the teacher's subjective perspective. 
By framing TOSS as an integrated procedural loop, this work complements recent evidence that humans interpret robot learning within a coherent inference framework \cite{hilpert_canyousee_2025}, allowing future research to conceptualize mental models in teaching as a dynamic and mutually adaptive process.

\subsection{Future Work \& Limitations}
Several constraints must be acknowledged.  
The evaluative Trigger dimensions may not necessarily be orthogonal and the features utilized by human teachers may overlap or correlate in ways not fully captured by the current qualitative analysis. Furthermore, this subjective ``evaluative space" may not map directly onto the technical feature space, necessitating further research. 
Second, the boundaries between Objectives
are not absolute. 
A Behavioral Objective for more precision 
may also serve a Goal Objective of error prevention. Rather than a set of rigid categories, they should be viewed as a fluid framework that helps disambiguate functional levels of subjective targets. More research is needed to further understand the nuances of teaching objectives in practice. 
Finally, 
while the rigid methodology and analysis provided rich, high-fidelity insights into teacher conceptualization of the teacher learner loop, the results remain an exploratory ``glimpse" into human pedagogical logic and serve as a foundational baseline rather than a definitive taxonomy. Future work must extend them through large-scale quantitative validation and closed-loop interactive studies to determine how they present when the teacher is faced with a live, adapting robot.

By accounting for the more nuanced and dynamic structure of the teaching \textit{process}, the TOSS framework provides a structured path for the development of more sophisticated computational teacher models. 
The results of this work could be combined with recent approaches that model variation in human feedback \cite{huangLeveragingVariationHuman2025} to integrate the subjective teacher perspective of 
TOSS with quantifiable  
formalizations to create dynamic 
simulations of human teaching, e.g as realistic oracles for algorithm evaluation, allowing research to move beyond simple signal matching.
If a robot can recognize 
a shift from a Coach to an Engineer strategy, it can adjust its interaction strategies (i.e. \cite{houGiveMeExample2024}) accordingly, thereby mitigating MMMs \cite{richterImprovingHumanRobotTeaching2025}, that lead to suboptimal interaction. 
This work also provides a direct 
query of teacher triggers and objectives, informing future analyses of teaching behavior.
Lastly, the structural flexibility revealed by TOSS necessitates a fundamental re-design of human-robot teaching settings. Future systems must move away from rigid, turn-taking interfaces, towards continuous, multi-level feedback integration, co-developed between teachers and robots \cite{habibianReviewCommunicatingRobot2023}. This requires robots that are not only receptive to procedural corrections but are also ``information-ready" to process declarative knowledge objectives or proactive environmental changes and integrate teacher flexibility to enable more natural human teaching for robots \cite{huangLeveragingVariationHuman2025}.
In line with previous work \cite{hilpertClosingTeacherLearnerLoop2024}, 
TOSS encourages the design of novel, teacher-adequate transparency mechanisms, compatible with the continuous, nuanced evaluations
informing the intervention.

\section{Conclusion}
This work presents the TOSS framework as an initial step to uncover the underlying logic that humans intuitively apply when making human-robot teaching decisions. By adopting a bottom-up, elicitation-based approach, this work shows that the human teacher's perspective on Human-Interactive Robot Learning is more than a reactive issuing of signals. Rather, it suggests that human teachers spontaneously adopt diverse roles, and within these roles prioritize different objectives, and consider different triggers and signals as relevant.

While only a first attempt, TOSS provides future research
with a theoretical foundation and an openly accessible dataset for understanding teaching decisions, simulating realistic oracles, and designing human-centered teaching settings and novel Human-Interactive Robot Learning algorithms that are needed for hybrid, synergistic collaboration.

\section*{Acknowledgements}
This research is sponsored by the Hybrid Intelligence project, grant number 024.004.022. Special thanks to Jonne Goedhart and Anna Lea Reinwarth for their support.

\begin{footnotesize}
\bibliographystyle{IEEEtran}
\bibliography{rlteach}

@article{braun_thematicanalysispractical_2021,
	title = {Thematic analysis: {A} practical guide},
	shorttitle = {Thematic analysis},
	url = {https://www.torrossa.com/it/resources/an/5282292},
	urldate = {2026-03-19},
	publisher = {SAGE publications Ltd},
	author = {Braun, Virginia and Clarke, Victoria},
	year = {2021},
}

@misc{gallouedecPandagymOpensourceGoalconditioned2021,
	title = {panda-gym: {Open}-source goal-conditioned environments for robotic learning},
	shorttitle = {panda-gym},
	url = {http://arxiv.org/abs/2106.13687},
	urldate = {2024-11-08},
	publisher = {arXiv},
	author = {Gallouédec, Quentin and Cazin, Nicolas and Dellandréa, Emmanuel and Chen, Liming},
	month = dec,
	year = {2021},
	note = {arXiv:2106.13687},
}

@misc{brockman_openaigym_2016,
	title = {{OpenAI} {Gym}},
	url = {http://arxiv.org/abs/1606.01540},
	doi = {10.48550/arXiv.1606.01540},
	urldate = {2026-02-25},
	publisher = {arXiv},
	author = {Brockman, Greg and Cheung, Vicki and Pettersson, Ludwig and Schneider, Jonas and Schulman, John and Tang, Jie and Zaremba, Wojciech},
	month = jun,
	year = {2016},
	note = {arXiv:1606.01540 [cs]},
}

@misc{hilpert_canyousee_2025,
	title = {Can you see how {I} learn? {Human} observers' inferences about {Reinforcement} {Learning} agents' learning processes},
	shorttitle = {Can you see how {I} learn?},
	url = {http://arxiv.org/abs/2506.13583},
	doi = {10.48550/arXiv.2506.13583},
	urldate = {2026-02-25},
	publisher = {arXiv},
	author = {Hilpert, Bernhard and Hou, Muhan and Baraka, Kim and Broekens, Joost},
	month = jun,
	year = {2025},
	note = {arXiv:2506.13583 [cs]},
}

@inproceedings{wesslerEmpiricalResearchAffective2021,
	title = {Empirical research in affective computing: {An} analysis of research practices and recommendations},
	shorttitle = {Empirical research in affective computing},
	url = {https://ieeexplore.ieee.org/abstract/document/9597418/},
	urldate = {2024-11-07},
	booktitle = {2021 9th {International} {Conference} on {Affective} {Computing} and {Intelligent} {Interaction} ({ACII})},
	publisher = {IEEE},
	author = {Wessler, Janet and Schneeberger, Tanja and Hilpert, Bernhard and Alles, Alexandra and Gebhard, Patrick},
	year = {2021},
	pages = {1--8},
}

@book{chernova_robotlearninghuman_2022,
	title = {Robot learning from human teachers},
	url = {https://books.google.com/books?hl=nl&lr=&id=s4RyEAAAQBAJ&oi=fnd&pg=PR1&dq=Chernova+%26+Thomaz,+2022+robot+learning+from+human+teachers&ots=ONerrKnmVW&sig=mUdU_Hsgz4zvGqzO-vy2hZPRJfc},
	urldate = {2026-02-25},
	publisher = {Springer Nature},
	author = {Chernova, Sonia and Thomaz, Andrea L.},
	year = {2022},
}

@article{baraka_humaninteractiverobotlearning_2026,
	title = {Human-{Interactive} {Robot} {Learning}: {Definition}, {Challenges}, and {Recommendations}},
	volume = {15},
	issn = {2573-9522},
	shorttitle = {Human-{Interactive} {Robot} {Learning}},
	url = {https://dl.acm.org/doi/10.1145/3779297},
	doi = {10.1145/3779297},
	language = {en},
	number = {2},
	urldate = {2026-02-25},
	journal = {ACM Transactions on Human-Robot Interaction},
	author = {Baraka, Kim and Idrees, Ifrah and Kessler Faulkner, Taylor and Biyik, Erdem and Booth, Serena and Chetouani, Mohamed and Grollman, Daniel H. and Saran, Akanksha and Senft, Emmanuel and Tulli, Silvia and Vollmer, Anna-Lisa and Andriella, Antonio and Beierling, Helen and Horter, Tiffany and Kober, Jens and Sheidlower, Isaac and Taylor, Matthew E. and Van Waveren, Sanne and Xiao, Xuesu},
	month = mar,
	year = {2026},
	pages = {1--31},
}

@misc{tulliInferringImplicitGoals2025,
	title = {Inferring {Implicit} {Goals} {Across} {Differing} {Task} {Models}},
	url = {http://arxiv.org/abs/2501.17704},
	doi = {10.48550/arXiv.2501.17704},
	urldate = {2025-02-11},
	publisher = {arXiv},
	author = {Tulli, Silvia and Vasileiou, Stylianos Loukas and Chetouani, Mohamed and Sreedharan, Sarath},
	month = jan,
	year = {2025},
	note = {arXiv:2501.17704 [cs]},
}

@misc{houGiveMeExample2024,
	title = {"{Give} {Me} an {Example} {Like} {This}": {Episodic} {Active} {Reinforcement} {Learning} from {Demonstrations}},
	shorttitle = {"{Give} {Me} an {Example} {Like} {This}"},
	url = {http://arxiv.org/abs/2406.03069},
	urldate = {2024-08-20},
	publisher = {arXiv},
	author = {Hou, Muhan and Hindriks, Koen and Eiben, A. E. and Baraka, Kim},
	month = jun,
	year = {2024},
	note = {arXiv:2406.03069 [cs]},
}

@inproceedings{faulknerInteractiveReinforcementLearning2020,
	title = {Interactive reinforcement learning with inaccurate feedback},
	url = {https://ieeexplore.ieee.org/abstract/document/9197219/},
	urldate = {2024-02-26},
	booktitle = {2020 {IEEE} {International} {Conference} on {Robotics} and {Automation} ({ICRA})},
	publisher = {IEEE},
	author = {Faulkner, Taylor A. Kessler and Short, Elaine Schaertl and Thomaz, Andrea L.},
	year = {2020},
	pages = {7498--7504},
}

@article{sarinPunishmentOrganizedPrinciples2021,
	title = {Punishment is organized around principles of communicative inference},
	volume = {208},
	url = {https://www.sciencedirect.com/science/article/pii/S0010027720303632},
	urldate = {2024-11-07},
	journal = {Cognition},
	publisher = {Elsevier},
	author = {Sarin, Arunima and Ho, Mark K. and Martin, Justin W. and Cushman, Fiery A.},
	year = {2021},
	pages = {104544},
}

@article{carrollUtilityLearningHumans2019,
	title = {On the utility of learning about humans for human-ai coordination},
	volume = {32},
	url = {https://proceedings.neurips.cc/paper/2019/hash/f5b1b89d98b7286673128a5fb112cb9a-Abstract.html},
	urldate = {2024-11-07},
	journal = {Advances in neural information processing systems},
	author = {Carroll, Micah and Shah, Rohin and Ho, Mark K. and Griffiths, Tom and Seshia, Sanjit and Abbeel, Pieter and Dragan, Anca},
	year = {2019},
}

@misc{gao_reinforcementlearningimperfect_2019,
	title = {Reinforcement {Learning} from {Imperfect} {Demonstrations}},
	url = {http://arxiv.org/abs/1802.05313},
	doi = {10.48550/arXiv.1802.05313},
	urldate = {2026-02-25},
	publisher = {arXiv},
	author = {Gao, Yang and Xu, Huazhe and Lin, Ji and Yu, Fisher and Levine, Sergey and Darrell, Trevor},
	month = may,
	year = {2019},
	note = {arXiv:1802.05313 [cs]},
}

@inproceedings{christofiHumanTeachingPatterns2025,
	address = {Eindhoven, Netherlands},
	title = {Human {Teaching} {Patterns} in {Interactive} {Robot} {Learning} from {Multiple} {Teaching} {Modalities}},
	copyright = {https://doi.org/10.15223/policy-029},
	isbn = {979-8-3315-8771-0},
	url = {https://ieeexplore.ieee.org/document/11217920/},
	doi = {10.1109/RO-MAN63969.2025.11217920},
	language = {en},
	urldate = {2026-02-20},
	booktitle = {2025 34th {IEEE} {International} {Conference} on {Robot} and {Human} {Interactive} {Communication} ({RO}-{MAN})},
	publisher = {IEEE},
	author = {Christofi, Konstantinos and Tichelaar, Caroline and Preciado, Daniel F. and Baraka, Kim},
	month = aug,
	year = {2025},
	pages = {1181--1187},
}

@inproceedings{ramaraj_understandingintentionshuman_2020,
	title = {Understanding intentions in human teaching to design interactive task learning robots},
	url = {https://raw.githubusercontent.com/SoarGroup/website-downloads/main/pubs/RSS_AI_ACR_2020-final.pdf},
	urldate = {2026-02-25},
	booktitle = {{RSS} 2020 {Workshop}: {AI} \& {Its} {Alternatives} in {Assistive} \& {Collaborative} {Robotics}: {Decoding} {Intent}},
	author = {Ramaraj, Preeti and Klenk, Matt and Mohan, Shiwali},
	year = {2020},
}

@inproceedings{ramaraj_unpackinghumanteachers_2021,
	address = {Vancouver, BC, Canada},
	title = {Unpacking {Human} {Teachers}’ {Intentions} for {Natural} {Interactive} {Task} {Learning}},
	copyright = {https://ieeexplore.ieee.org/Xplorehelp/downloads/license-information/IEEE.html},
	isbn = {978-1-6654-0492-1},
	url = {https://ieeexplore.ieee.org/document/9515448/},
	doi = {10.1109/RO-MAN50785.2021.9515448},
	language = {en},
	urldate = {2026-02-25},
	booktitle = {2021 30th {IEEE} {International} {Conference} on {Robot} \& {Human} {Interactive} {Communication} ({RO}-{MAN})},
	publisher = {IEEE},
	author = {Ramaraj, Preeti and Ortiz, Charles L. and Mohan, Shiwali},
	month = aug,
	year = {2021},
	pages = {1173--1180},
}

@misc{villareale_understandingmentalmodels_2021,
	title = {Understanding {Mental} {Models} of {AI} through {Player}-{AI} {Interaction}},
	url = {http://arxiv.org/abs/2103.16168},
	doi = {10.48550/arXiv.2103.16168},
	urldate = {2026-02-25},
	publisher = {arXiv},
	author = {Villareale, Jennifer and Zhu, Jichen},
	month = mar,
	year = {2021},
	note = {arXiv:2103.16168 [cs]},
}

@inproceedings{huangEffectRobotErrors2024,
	title = {On the {Effect} of {Robot} {Errors} on {Human} {Teaching} {Dynamics}},
	url = {http://arxiv.org/abs/2409.09827},
	doi = {10.1145/3687272.3688320},
	language = {en},
	urldate = {2026-02-20},
	booktitle = {Proceedings of the 12th {International} {Conference} on {Human}-{Agent} {Interaction}},
	author = {Huang, Jindan and Sheidlower, Isaac and Aronson, Reuben M. and Short, Elaine Schaertl},
	month = nov,
	year = {2024},
	note = {arXiv:2409.09827 [cs]},
	pages = {150--159},
}

@article{loftinLearningBehaviorsHumandelivered2016,
	title = {Learning behaviors via human-delivered discrete feedback: modeling implicit feedback strategies to speed up learning},
	volume = {30},
	issn = {1387-2532, 1573-7454},
	shorttitle = {Learning behaviors via human-delivered discrete feedback},
	url = {http://link.springer.com/10.1007/s10458-015-9283-7},
	doi = {10.1007/s10458-015-9283-7},
	language = {en},
	number = {1},
	urldate = {2026-02-20},
	journal = {Autonomous Agents and Multi-Agent Systems},
	author = {Loftin, Robert and Peng, Bei and MacGlashan, James and Littman, Michael L. and Taylor, Matthew E. and Huang, Jeff and Roberts, David L.},
	month = jan,
	year = {2016},
	pages = {30--59},
}

@inproceedings{huangLeveragingVariationHuman2025,
	address = {Melbourne, Australia},
	title = {Leveraging {Variation} in {Human} {Feedback} {Expression} to {Enable} {Natural} {Human} {Teaching} for {Robots}},
	copyright = {https://doi.org/10.15223/policy-029},
	isbn = {979-8-3503-7893-1},
	url = {https://ieeexplore.ieee.org/document/10974126/},
	doi = {10.1109/HRI61500.2025.10974126},
	language = {en},
	urldate = {2026-02-20},
	booktitle = {2025 20th {ACM}/{IEEE} {International} {Conference} on {Human}-{Robot} {Interaction} ({HRI})},
	publisher = {IEEE},
	author = {Huang, Jindan},
	month = mar,
	year = {2025},
	pages = {1851--1853},
}

@inproceedings{fangCHARMConsideringHuman2025,
	address = {Eindhoven, Netherlands},
	title = {{CHARM}: {Considering} {Human} {Attributes} for {Reinforcement} {Modeling}},
	copyright = {https://doi.org/10.15223/policy-029},
	isbn = {979-8-3315-8771-0},
	shorttitle = {{CHARM}},
	url = {https://ieeexplore.ieee.org/document/11217604/},
	doi = {10.1109/RO-MAN63969.2025.11217604},
	language = {en},
	urldate = {2026-02-20},
	booktitle = {2025 34th {IEEE} {International} {Conference} on {Robot} and {Human} {Interactive} {Communication} ({RO}-{MAN})},
	publisher = {IEEE},
	author = {Fang, Qidi and Yu, Hang and Fang, Shijie and Huang, Jindan and Chen, Qiuyu and Aronson, Reuben M. and Short, Elaine S.},
	month = aug,
	year = {2025},
	pages = {7--14},
}

@inproceedings{hilpertClosingTeacherLearnerLoop2024,
	address = {Glasgow, United Kingdom},
	title = {Closing the {Teacher}-{Learner} {Loop}: {The} {Role} of {Affective} {Signals} in {Interactive} {RL}},
	copyright = {https://doi.org/10.15223/policy-029},
	isbn = {979-8-3315-1645-1},
	shorttitle = {Closing the {Teacher}-{Learner} {Loop}},
	url = {https://ieeexplore.ieee.org/document/10970360/},
	doi = {10.1109/ACIIW63320.2024.00020},
	language = {en},
	urldate = {2026-02-18},
	booktitle = {2024 12th {International} {Conference} on {Affective} {Computing} and {Intelligent} {Interaction} {Workshops} and {Demos} ({ACIIW})},
	publisher = {IEEE},
	author = {Hilpert, Bernhard},
	month = sep,
	year = {2024},
	pages = {97--101},
}

@misc{richterImprovingHumanRobotTeaching2025,
	title = {Improving {Human}-{Robot} {Teaching} by {Quantifying} and {Reducing} {Mental} {Model} {Mismatch}},
	url = {http://arxiv.org/abs/2501.04755},
	doi = {10.48550/arXiv.2501.04755},
	urldate = {2025-01-14},
	publisher = {arXiv},
	author = {Richter, Phillip and Wersing, Heiko and Vollmer, Anna-Lisa},
	month = jan,
	year = {2025},
	note = {arXiv:2501.04755 [cs]},
}

@inproceedings{guillorySimultaneousLearningCovering2011,
	title = {Simultaneous learning and covering with adversarial noise},
	url = {https://icml.cc/2011/papers/261_icmlpaper.pdf},
	urldate = {2025-02-10},
	booktitle = {Proceedings of the 28th {International} {Conference} on {Machine} {Learning} ({ICML}-11)},
	author = {Guillory, Andrew and Bilmes, Jeff A.},
	year = {2011},
	pages = {369--376},
}

@inproceedings{bansalAccuracyRoleMental2019,
	title = {Beyond accuracy: {The} role of mental models in human-{AI} team performance},
	volume = {7},
	shorttitle = {Beyond accuracy},
	url = {https://aaai.org/ojs/index.php/HCOMP/article/view/5285},
	urldate = {2025-01-15},
	booktitle = {Proceedings of the {AAAI} conference on human computation and crowdsourcing},
	author = {Bansal, Gagan and Nushi, Besmira and Kamar, Ece and Lasecki, Walter S. and Weld, Daniel S. and Horvitz, Eric},
	year = {2019},
	pages = {2--11},
}

@inproceedings{houShapingImbalanceBalance2023,
	title = {Shaping {Imbalance} into {Balance}: {Active} {Robot} {Guidance} of {Human} {Teachers} for {Better} {Learning} from {Demonstrations}},
	shorttitle = {Shaping {Imbalance} into {Balance}},
	url = {https://ieeexplore.ieee.org/abstract/document/10309481/},
	urldate = {2024-05-21},
	booktitle = {2023 32nd {IEEE} {International} {Conference} on {Robot} and {Human} {Interactive} {Communication} ({RO}-{MAN})},
	publisher = {IEEE},
	author = {Hou, Muhan and Hindriks, Koen and Eiben, A. E. and Baraka, Kim},
	year = {2023},
	pages = {1737--1744},
}

@inproceedings{houProcessOrientedFrameworkRobot2023,
	title = {A {Process}-{Oriented} {Framework} for {Robot} {Imitation} {Learning} in {Human}-{Centered} {Interactive} {Tasks}},
	url = {https://ieeexplore.ieee.org/abstract/document/10309326/},
	urldate = {2024-05-21},
	booktitle = {2023 32nd {IEEE} {International} {Conference} on {Robot} and {Human} {Interactive} {Communication} ({RO}-{MAN})},
	publisher = {IEEE},
	author = {Hou, Muhan and Hindriks, Koen and Eiben, A. E. and Baraka, Kim},
	year = {2023},
	pages = {1745--1752},
}

@article{hoPeopleTeachRewards2019,
	title = {People teach with rewards and punishments as communication, not reinforcements.},
	volume = {148},
	url = {https://psycnet.apa.org/journals/xge/148/3/520/},
	number = {3},
	urldate = {2024-02-26},
	journal = {Journal of Experimental Psychology: General},
	publisher = {American Psychological Association},
	author = {Ho, Mark K. and Cushman, Fiery and Littman, Michael L. and Austerweil, Joseph L.},
	year = {2019},
	pages = {520},
}

@incollection{chetouaniInteractiveRobotLearning2023,
	address = {Cham},
	title = {Interactive {Robot} {Learning}: {An} {Overview}},
	volume = {13500},
	isbn = {978-3-031-24348-6 978-3-031-24349-3},
	shorttitle = {Interactive {Robot} {Learning}},
	url = {https://link.springer.com/10.1007/978-3-031-24349-3_9},
	doi = {10.1007/978-3-031-24349-3_9},
	language = {en},
	urldate = {2024-02-26},
	booktitle = {Human-{Centered} {Artificial} {Intelligence}},
	publisher = {Springer International Publishing},
	author = {Chetouani, Mohamed},
	editor = {Chetouani, Mohamed and Dignum, Virginia and Lukowicz, Paul and Sierra, Carles},
	year = {2023},
	note = {Series Title: Lecture Notes in Computer Science},
	pages = {140--172},
}

@article{habibianReviewCommunicatingRobot2023,
	title = {A review of communicating robot learning during human-robot interaction},
	url = {https://collab.me.vt.edu/pdfs/soheil_ijrr2023.pdf},
	urldate = {2024-10-15},
	journal = {arXiv preprint arXiv:2312.00948},
	author = {Habibian, Soheil and Valdivia, Antonio Alvarez and Blumenschein, Laura H. and Losey, Dylan P.},
	year = {2023},
}

@article{akataResearchAgendaHybrid2020,
	title = {A research agenda for hybrid intelligence: augmenting human intellect with collaborative, adaptive, responsible, and explainable artificial intelligence},
	volume = {53},
	shorttitle = {A research agenda for hybrid intelligence},
	url = {https://ieeexplore.ieee.org/abstract/document/9153877/},
	number = {8},
	urldate = {2024-06-10},
	journal = {Computer},
	publisher = {IEEE},
	author = {Akata, Zeynep and Balliet, Dan and De Rijke, Maarten and Dignum, Frank and Dignum, Virginia and Eiben, Guszti and Fokkens, Antske and Grossi, Davide and Hindriks, Koen and Hoos, Holger},
	year = {2020},
	pages = {18--28},
}
\end{footnotesize}

\end{document}